\pdfoutput=1
\documentclass[11pt]{article}

\usepackage[final]{acl}

\usepackage{times}
\usepackage{latexsym}

\usepackage[T1]{fontenc}
\usepackage[utf8]{inputenc}

\usepackage{microtype}

\usepackage{inconsolata}

\usepackage{graphicx}
\usepackage{subcaption}
\usepackage{tabularx}
\usepackage{booktabs}

\title{Why Do Self-Harm Prediction Models Struggle to Generalise? \\Lexical and Semantic Variations in Emergency Department Triage Notes\thanks{This paper has been accepted to the Workshop on Computational Linguistics and Clinical Psychology (CLPsych 2026).}}

\author{
 \textbf{Liuliu Chen\textsuperscript{1}},
 \textbf{Mike Conway\textsuperscript{1}},
\textbf{Jo Robinson\textsuperscript{2,3}},
\textbf{Vlada Rozova\textsuperscript{1,4}}
\\
\\
 \textsuperscript{1}School of Computing and Information Systems, The University of Melbourne, Australia
 \\
\textsuperscript{2}Orygen, The National Centre of Excellence in Youth Mental Health, Australia
\\
 \textsuperscript{3}Centre for Youth Mental Health, The University of Melbourne, Australia
 \\
 \textsuperscript{4}Centre for Digital Transformation of Health, The University of Melbourne, Australia
\\
 \small{
\textbf{Correspondence:} \href{mailto:liuliuc@student.unimelb.edu.au}{liuliuc@student.unimelb.edu.au}
 }
}

\begin{document}
\maketitle
\begin{abstract}
% Prediction model generalisability is challenging in the clinical domain. To explore what factors might impact model generalization in self-harm prediction across hospitals, we compared two emergency department triage note datasets by analyzing lexical characteristics, highly associated predictive features, and salient topics. Our results indicate the variation in lexical choices and feature importance in predicting self-harm across hospitals. These findings revealed potential reasons why models struggle to generalize and contribute to understanding cross-hospital model adaptation. They also highlight potential methods to improve model generalisability.

Self-harm presentations to emergency departments (EDs) are strongly associated with higher suicide risk. NLP models have shown robust performance in detecting self-harm from triage notes within single hospitals, yet performance often declines across institutions. To examine potential causes, we compare ED triage notes from two hospitals by analyzing lexical characteristics, highly associated predictive features, and salient topics.
Our results reveal variation in lexical expression and feature importance related to self-harm across hospitals, despite consistent core themes such as self-poisoning and self-injury. These documentation differences are associated with reduced cross-site performance. Our findings provide insight into how institutional variation affects the identification of self-harm in clinical text and highlight potential methods to improve model generalisability.
\end{abstract}

\section{Introduction}

Generalisability of predictive models is important in the clinical domain, where data collection and annotation are often expensive and logistically challenging \cite{goetz2024generalization}. Yet achieving robust generalisation remains a key challenge in real-world clinical applications for various reasons. For example, limited representativeness in the training data can impact model transfer to new datasets \cite{goetz2024generalization}, while the predictive factors that support or hinder generalisability are poorly understood \cite{futoma2021generalization}. 
Prior work on clinical NLP portability has also shown that model performance can degrade substantially when systems are applied to new clinical domains, motivating domain adaptation strategies that account for differences in documentation style, institution, specialty, and data-sharing constraints \cite{laparra2020rethinking}.

% ome predictive models overfit the training data for better performance
% , further raising concerns about the consistency of model performance beyond the development environment

Self-harm -- defined as ``an intentional act of self-poisoning (e.g., drug overdose) or self-injury (e.g., self-cutting) regardless of suicidal intent" -- has become increasingly common among young people and is a major global health concern \cite{witt2023characteristics}. This has driven researchers to develop methods for self-harm detection in clinical documentation over the years \cite{obeid2020identifying, ayre2021developing, iorfino2020predicting, rozova2022detection}. Recently published studies using natural language processing have shown strong performance. For instance, \citet{obeid2020identifying} reported an AUROC of 0.88 and an F1-score of 0.77 on electronic health records (EHRs). \citet{rozova2022detection} reported an AUPRC of 0.85 when applying a Gradient Boosting model to emergency department (ED) triage notes. However, these results were obtained on test sets with the same distributions as the training data. Whether these models can generalise to other contexts remains under-explored. 

Building on \citeposs{rozova2022detection} study, we aim to examine the model's external validity in a different hospital context. The model was trained on ED data from the Royal Melbourne Hospital (RMH), a major tertiary referral centre in Melbourne, Victoria, Australia. When applied to Latrobe Regional Health (LRH), a regional hospital serving rural communities in Victoria, model performance declined, with AUPRC decreasing from 0.85 $\pm$ 0.01 to 0.78 $\pm$ 0.01. To investigate the potential contributors to this cross-site generalisation gap, we conduct a corpora comparison analysis on ED triage notes between RMH and LRH.

Triage notes describe the reason for patient presentation to the ED and are typically short and hastily written, resulting in non-standard grammar, misspellings and extensive use of abbreviations, which can vary considerably between hospital systems. %This introduces additional variability to the language of triage notes. 
Through the analysis of lexical characteristics, features highly associated with the outcome variable, and topic modelling, we aim to identify potential contributors to the model's diminished performance when ported to a different context.

\section{Methodology}
\subsection{Datasets}
This study uses ED triage notes from RMH and LRH between 2012 and 2017 \cite{witt2023characteristics}, which were manually annotated by psychology experts in suicide prevention as positive (\textbf{Self-harm}) or negative (\textbf{Control}). The RMH dataset consisted of N=399,111 notes with 1.4\% notes annotated as positive. The LRH dataset contained N=171,170 notes, of which 1.7\% were considered positive. 

RMH is a large metropolitan public tertiary hospital, while LRH is the principal referral hospital for the Gippsland region, serving a predominantly rural and regional catchment. Table~\ref{tab:hospital_comparison} summarises key contextual differences between the two hospitals.

\setlength{\tabcolsep}{1mm}
\begin{table}[t]
\centering
\small
\caption{Comparisons between RMH and LRH.}
\label{tab:hospital_comparison}
\begin{tabularx}{\linewidth}{XXX}
\toprule
 & \textbf{RMH} & \textbf{LRH} \\
\midrule
Funding & Public & Public \\
Location & Metropolitan & Regional \\
ED records, 2012--2017 & 399{,}111 & 171{,}170 \\
Catchment area & 137 sq. km & 42{,}000 sq. km \\
Population served & 550{,}000 & 300{,}000 \\
Emergency mental health team & Yes & No \\
Age & 48$\pm$21 & 47$\pm$23 \\
Sex & & \\
\quad Female & 48\% & 52\% \\
\quad Male & 52\% & 48\% \\
\quad Intersex & $<1\%$ & $<1\%$ \\
\quad Unknown & $<1\%$ & $<1\%$ \\
\bottomrule
\end{tabularx}
\end{table}

% \begin{table}[ht]
% \centering
% \small
% \begin{tabular}{lllll}
% \hline
%     & SH   & Controls & Total  & Duration \\ \hline
% RMH & 5408 & 393,703  & 399,111 & 2012-2017 \\
% LRH & 2883 & 168,287  & 171,170 & 2012-2017 \\ \hline
% \end{tabular}
% \caption{Overview number of RMH and LRH datasets.}
% \label{tab: dataset_overview}
% \end{table}

\subsection{Corpora Comparison}
To compare the corpora, we examined lexical characteristics (e.g., number of characters and sentences), identified important features in each dataset, and applied topic modelling to explore the most frequent topics in the self-harm group.

\textbf{Important Features Selection \space}
Following \citet{rozova2022detection}, we employed TF–IDF representations to ensure comparability and enable external validation. We computed the TF-IDF matrix and selected important features using both the Chi-Square and XGBoost algorithms. Chi-Square measures the correlation between a feature and the target variable (i.e., self-harm), and features with \textit{p} < .001 were retained. XGBoost feature importance reflects each feature’s predictive contribution, and we selected features above the 95th percentile (top 5\%). We applied Chi-Square test and XGBoost independently on each dataset, both across all years and separately for each year. The final set of important features was the intersection of features selected by both methods. We further compare these final selected important features between the two hospitals.

\textbf{Topic Modelling \space}
To explore common themes in the self-harm class, BERTopic was applied \cite{grootendorst2022bertopic} separately to each dataset, using only the positive class. We selected `all-mpnet-base-v2' as our sentence embedding model, which currently performs the best in capturing semantic information \cite{sbertPretrainedModels}. The topic-wise embedding semantic similarity between hospitals was calculated by cosine similarity. To identify unique topics, we set a threshold of 0.75, selected based on a sensitivity analysis across the 0.60--0.90 range.

\section{Results}
\subsection{Lexical Characteristics}
\begin{figure}[t]
    \centering
    \begin{subfigure}[b]{0.49\linewidth} % Fix syntax
        \centering
        \includegraphics[width=\linewidth]{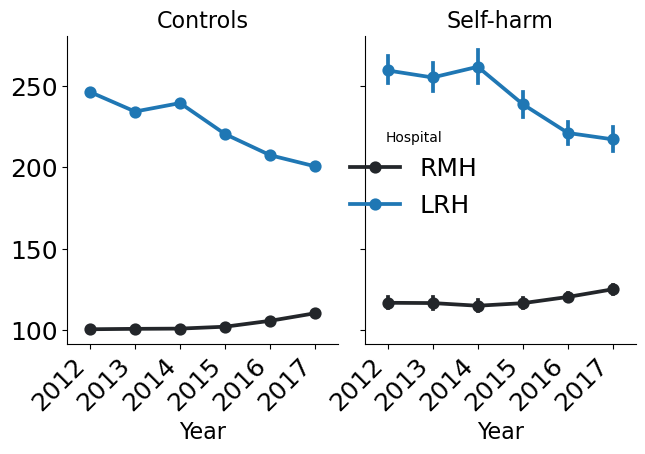}
        \caption{Character count}
        \label{fig:char_count}
    \end{subfigure}
    \hfill
    \begin{subfigure}[b]{0.49\linewidth} % Fix syntax
        \centering
        \includegraphics[width=\textwidth]{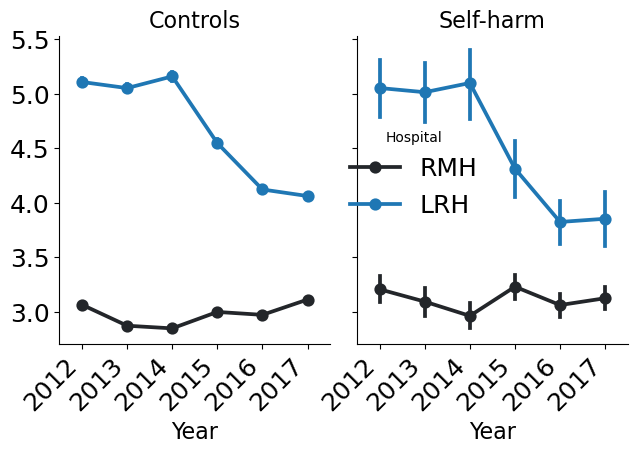}
        \caption{Sentence count}
        \label{fig:sent_count}
    \end{subfigure}
    \caption{Lexical characteristics}
    \label{fig:meta_ft_plot}
\end{figure}

Figure \ref{fig:meta_ft_plot} provides descriptive statistics regarding the number of characters and sentences. 

\subsection{Comparison of Important Features}
\subsubsection{Overall Comparison}
Across the entire dataset, we identified 472 unigram, 701 bigram, and 637 trigram important features at RMH and 365 unigram, 460 bigram, and 426 trigram features at LRH. The top 10 selected features for each hospital are presented in Appendix, Table \ref{tab:fts_overall_comparison}.

Of those, 213 unigram, 156 bigram, and 136 trigram features were shared between the two datasets, indicating their transferability. Table \ref{tab: shared_fts} presents the 10 most important shared features. As shown, these features mostly contain generic self-harm related keywords (e.g., \textit{self harm}, \textit{self inflicted}), self-harm methods (predominantly self-poisoning, e.g., \textit{od}, \textit{intentional od}), and self-injury (e.g., \textit{superficial cuts to}). Suicide-related keywords are also common in shared features, such as \textit{suicidal}, \textit{suicide}, and \textit{suicide intent}.

\setlength{\tabcolsep}{1mm}
\begin{table}[t]
\centering
\small
    \begin{tabularx}{\linewidth}{p{1.1cm} X}
    \hline
            & \textbf{Shared features}                                                                                                                                                                             \\ \hline
    \textbf{Unigram} & od, pain, self, tablets, overdose, sob, resolved, suicidal, suicide, attempt                                                                                                             \\
    \textbf{Bigram}  & self inflicted, od of, self harm, intentional od, suicide attempt, to end, to kill, polypharmacy od, wanted to, suicide intent                                                         \\
    \textbf{Trigram} & superficial lac(s) to, superficial cuts to, self harm to, self harm lacs, states has taken, with razor blade, with suicidal intent, to kill self, wants to die, pt has taken \\ \hline
    \end{tabularx}
\caption{Top 10 features identified in both RMH and LRH, sorted by average XGBoost importance ranking.}
\label{tab: shared_fts}
\end{table}

\begin{figure}[h]
    \centering
  \includegraphics[width=.9\linewidth]{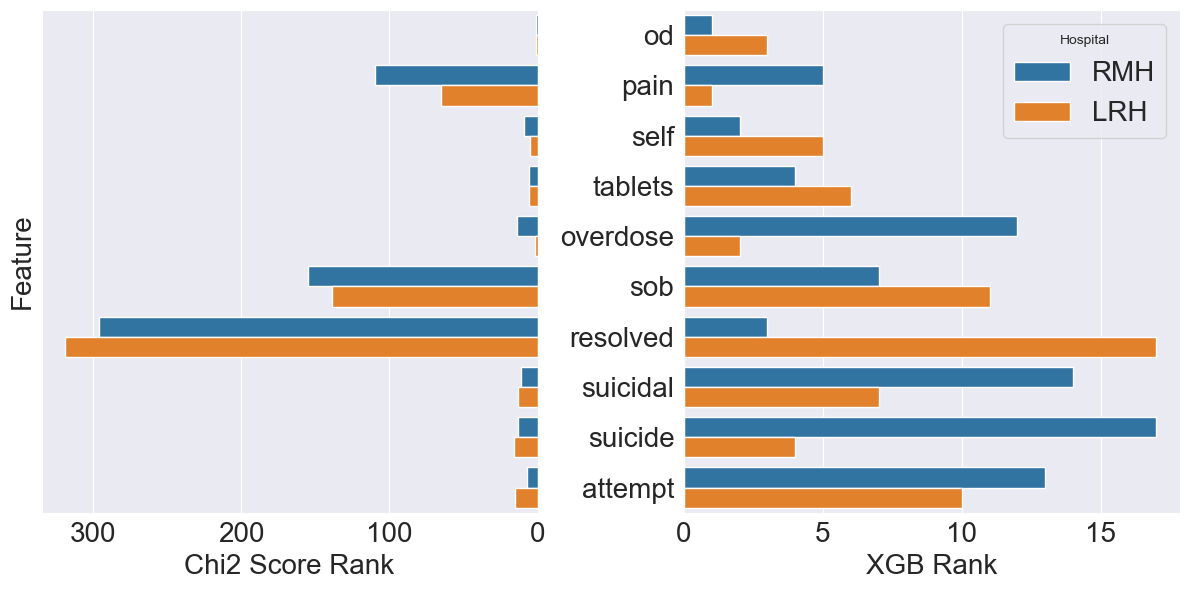}
  \caption{Ranking of shared unigram features, sorted by average XGBoost importance score ranking.}
  \label{fig:shared_ft_xgb}
\end{figure}

% [Differences in same feature -chi2/xgboost ranking]
We plotted the ranking of shared unigram features in Figure \ref{fig:shared_ft_xgb}. Although present in both hospitals, these features differed in their statistical association with self-harm (Chi-Square) and predictive importance (XGBoost). Additionally, statistical association does not always indicate predictive power for classification. For example, \textit{suicide} in RMH ranks highly based on the Chi-Square score but does not necessarily hold the highest predictive importance in model predictions.

% [Differences in same feature - tf-idf representation]
We compared the TF-IDF representations for each hospital. While feature selection was previously performed independently on each dataset, the TF-IDF representation here was computed using a shared vocabulary derived from both datasets to enable direct comparison.

\begin{figure}[t]
    \centering
    % \begin{subfigure}[b]{0.325\linewidth}
    %     \centering
    %     \includegraphics[width=\linewidth]{latex/plots/dist_tfidf_attempt.png}
    %     \caption{\textit{Attempt}}
    % \end{subfigure}
    % \hfill
    \begin{subfigure}[b]{0.47\linewidth}
        \centering
        \includegraphics[width=\linewidth]{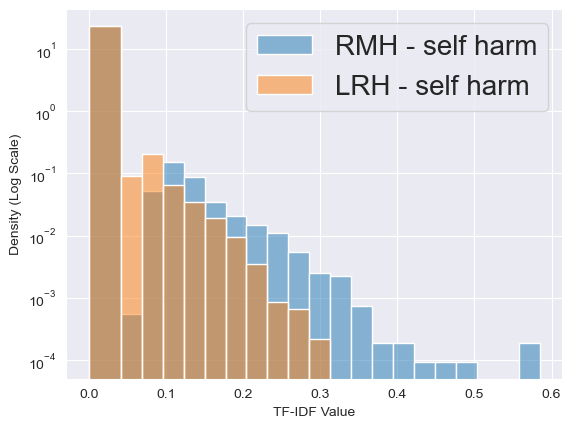}
        \caption{\textit{Self harm}}
        \label{}
    \end{subfigure}
    \hfill
    \begin{subfigure}[b]{0.47\linewidth}
        \centering
        \includegraphics[width=\linewidth]{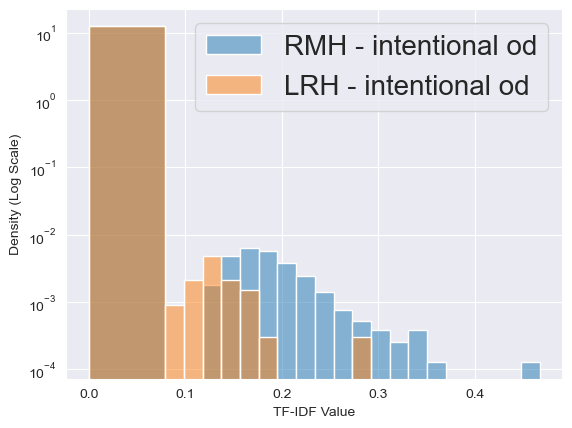}
        \caption{\textit{Intentional od}}
        \label{}
    \end{subfigure}
    \caption{Distribution of TF-IDF values}
    \label{fig:dist_tf_idf}
\end{figure}

% Table \ref{tab:cos_sim_tfidf} presents the overall similarity between the TF-IDF vectors of RMH and LRH. 
Overall, the cosine similarity between site-level mean TF-IDF vectors for RMH and LRH was 0.822 for unigrams, 0.628 for bigrams, and 0.473 for trigrams.
For the top 10 shared features listed in Table \ref{tab: shared_fts}, we further compared the distributions of document-level TF-IDF values between hospitals. All unigram features differed significantly between the two hospitals (\textit{p} < .01 for \textit{resolved}, \textit{p} < .001 for all others). Among bigram features, TF-IDF values for \textit{suicide intent} and \textit{suicide attempt} were similar, while the remaining bigram features showed significant differences (\textit{p} < .01 for \textit{self-inflicted}, \textit{p} < .001 for all others). All trigram features differed significantly between hospitals (\textit{p} < .001). Figure \ref{fig:dist_tf_idf} shows examples of different distributions of TF-IDF values for terms \textit{self harm} and \textit{intentional od} between hospitals.

% \begin{table}[ht]
%     \centering
%     \small
%     \begin{tabular}{cccc}
%     \hline
%        & \textbf{Unigram} & \textbf{Bigram} & \textbf{Trigram} \\ \hline
%        \textbf{Cos Similarity}  & 0.8219 & 0.6279 & 0.4725 \\
%        \hline
%     \end{tabular}
%     \caption{Cosine similarity on whole TF-IDF vectors between RMH and LRH}
%     \label{tab:cos_sim_tfidf}
% \end{table}

% [Differences in feature names]
Uniquely important features were also identified in each dataset. Some were misspellings (e.g., \textit{paracetOmol}) or dosage values (e.g., \textit{55mg}), which were excluded to focus on features with potentially meaningful insights, as shown in Table \ref{tab: unique_features}. Compared to RMH, LRH appears to have more police involvement in triage notes, with terms such as \textit{with police} and \textit{section 351} (Victorian Mental Health Act - apprehension of a person by police). In contrast, RMH contains more features related to social and psychological stressors, including \textit{recent relationship breakdown}, \textit{social stressors}.

\setlength{\tabcolsep}{1mm}
\begin{table}[t]
\centering
\small
\begin{tabularx}{\linewidth}{p{1.1cm} p{3.1cm} X}
\hline
    & \textbf{RMH}                                            & \textbf{LRH}     \\  \hline 
\textbf{Unigram} & jump, dizzy, bpd, dementia, stab, life, changes, xanax                                                                                                                               & biba, dose, bandaged, tied, children, battery, community, droop                                                                                                      \\
\textbf{Bigram}  & sudden onset, social stressors, anxiety depression, attempt to, phx depression, flank pain,  with etoh                                                                         & on palpation, biba after, with police, analgesia taken, dose of, with nil, attempted od, alleged overdose                                                                 \\
\textbf{Trigram} & recent relationship breakdown, previous self harm, recent social stressors, drowsy at triage, with stanley knife & to left forearm, under section 351, under section 10, with self inflicted, biba with police\\
\hline
\end{tabularx}
\caption{Features selected only in RMH or LRH.}
\label{tab: unique_features}
\end{table}

\subsubsection{Longitudinal Analysis}
We further applied feature selection independently to each year and each hospital (Appendix, Figure \ref{fig:total_ft_count}). Some features remained consistent across all years in both datasets (unigram: 20, bigram: 9, trigram: 2). These features are primarily related to self-poisoning (e.g., \textit{OD, overdose, tablets}), suicide (e.g., \textit{suicidal, suicide attempt}), and self-inflicted harm (e.g., \textit{inflicted, forearm}). 
% Additionally, \textit{depression} and \textit{diazepam} were consistently identified as key features across the years in both hospitals.

\begin{figure*}[t]
\begin{subfigure}[b]{0.25\linewidth}
        \centering
        \includegraphics[width=\linewidth]{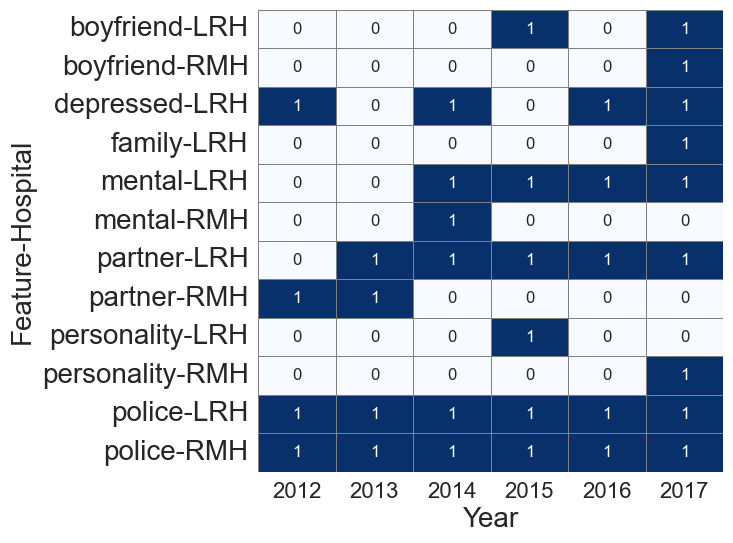}
        \caption{Social factors}
    \end{subfigure}
    \hfill
\begin{subfigure}[b]{0.25\linewidth}
        \centering
        \includegraphics[width=\linewidth]{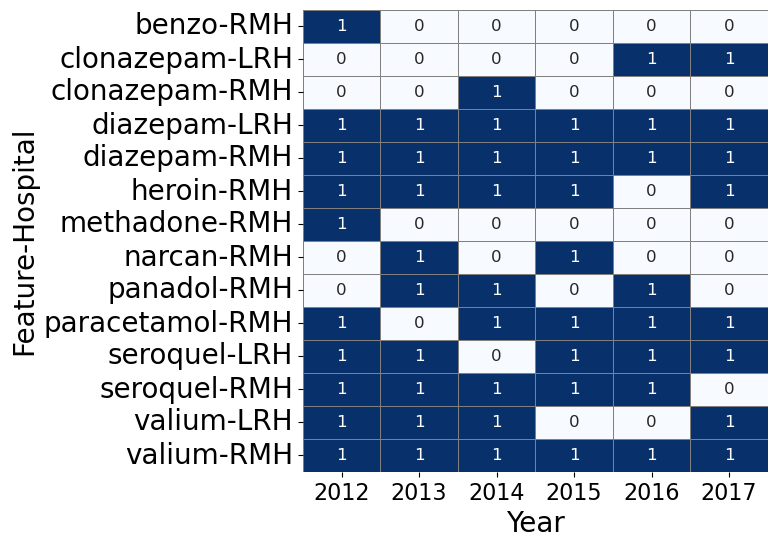}
        \caption{Medication}
    \end{subfigure}
     \hfill
\begin{subfigure}[b]{0.23\linewidth}
        \centering
        \includegraphics[width=\linewidth]{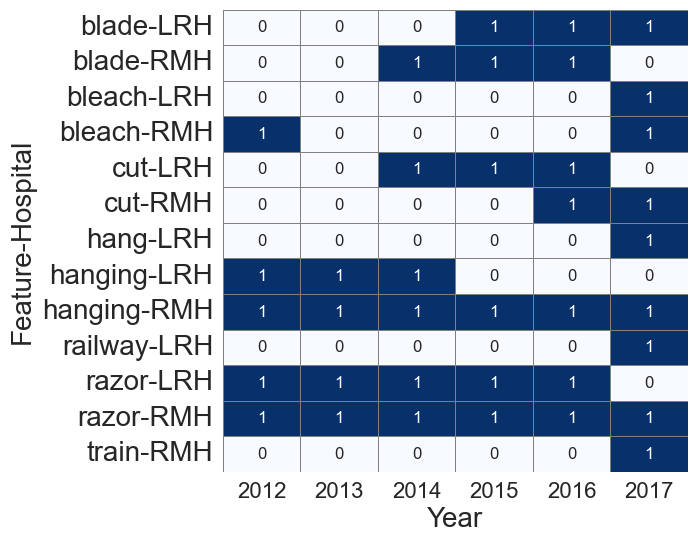}
        \caption{Self-harm methods}
    \end{subfigure}
     \hfill
\begin{subfigure}[b]{0.24\linewidth}
        \centering
        \includegraphics[width=\linewidth]{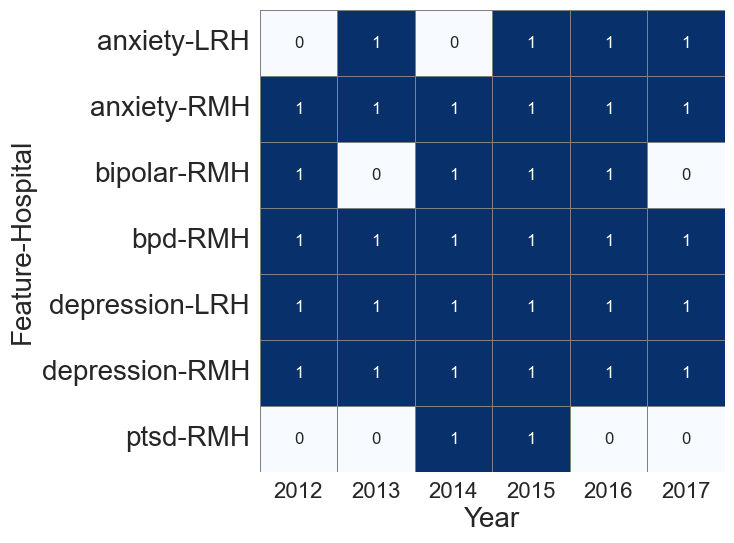}
        \caption{Mental health disorders}
    \end{subfigure}
  \caption{Heatmap of feature presence for feature-hospital across years (1: present, 0: absent). Note: some features appear in only one hospital because features absent in all years for a given hospital were removed.}
\label{fig: heatmap}
\end{figure*}

Focusing on social and mental factors, self-harm methods, mental disorders, and medication, we examined the consistency of related unigram features, as shown in Figure \ref{fig: heatmap}. Certain word features, such as \textit{police}, \textit{diazepam}, and \textit{depression}, appear consistently across years in both hospitals. Social factors like \textit{partner} are more consistent in RMH, as well as suicide methods (e.g., \textit{hanging/hang}). Medication terms also show some variations between hospitals. For example, \textit{paracetamol} was present in most years in RMH but was not selected in LRH. Regarding mental health disorders, most related terms appear consistently in RMH, whereas some features, such as \textit{bipolar}, were not selected in any year for LRH.

\subsection{Topic Modelling}
BERTopic models identified 30 topics in LRH and 74 in RMH. After manually reviewing, we found that most were related to self-harm methods, primarily overdosing on specific medications and dosages. Other self-harm methods also emerged, such as self-inflicted injuries and hanging.

The average semantic similarity across all documents was 0.46, while the overall topic representation (top-10 keywords) had a similarity score of 0.68. To further analyse topic overlap, we calculated pairwise topic semantic similarity and identified unique topics in each hospital. Based on sensitivity analysis across the 0.60--0.90 cosine similarity range, LRH consistently showed no unique topics across thresholds of 0.60--0.85, while the number of unique RMH topics varied. We therefore used 0.75 as a heuristic threshold, as it provided an intermediate level of strictness within the tested range.

We found 10 unique topics in RMH, while no unique topics were identified in LRH. Table \ref{tab: bertopic_rmh} presents the top 5 unique topics in RMH. Topics 42 and 56 are associated with overdose and poisoning presentations. Topics 51 and 14 primarily reflect psychiatric complexity, such as \textit{psychosis} and \textit{BPD}. Topic 27 captures hanging-related presentations.

Interestingly, while semantic embeddings of topics showed high similarity, their Jaccard similarity %(word overlap) 
was much lower. This suggests that while both hospitals discuss similar topics, the specific lexicons used may differ.

\setlength{\tabcolsep}{1mm}
\begin{table}[t]
\centering
\small
\begin{tabularx}{\linewidth}{p{0.5cm} X}
\hline
\textbf{No.} & \textbf{Representative words} \\ \hline
14                 & etoh, valium, valiumx4, bpd, suicidal, anxiety, 22mg, 5mg, abuse, harm                             \\
27                 & rope, hung, hanging, hang, distress, neck, haematoma, fell, seizure, spine                              \\
42                 & olanzipine, 80mg, 10mg, benztropine, 4mg, fluoxetine, 5mg, 30mg, polypharmacy, haloperidolol         \\
51                 & psychosis, meds, anxiety, pmhx, stressors, midazolam, discharged, abusive, bandaged, 351                                           \\
56                 & suicidal, suicide, poisoning, cpr, monoxide, unconscious, hypoxic, triaged, lethargy, catatonic \\
\hline
\end{tabularx}
\caption{BERTopic identified unique topics in RMH}
\label{tab: bertopic_rmh}
\end{table}

\section{Discussion}
The above analysis highlights key differences between two hospitals, offering insights into why the TF-IDF Chi-Square/XGBoost model from \citet{rozova2022detection} performed less effectively at LRH.

\textbf{Variations in TF-IDF lexical features \space}
TF-IDF analysis revealed that while certain features are transferable between hospitals, their associations and relative importance differ, and each hospital displayed unique features. Mental health–related and social factor terms were more characteristic of RMH than LRH, and the selected medication types also differed. This suggests that term frequency distributions across hospitals vary, potentially due to differences in documentation styles or the geodemographics of patient populations. For instance, RMH (but not LRH) has an ED mental health team specifically trained to assess mental health status. These findings suggest that models trained within a single site may partially rely on site-specific documentation patterns rather than consistently capturing stable indicators of self-harm risk.

\textbf{Semantic similarity with lexical variation in BERTopic \space} 
BERTopic analysis showed high semantic similarity across most detected topics, indicating the thematic consistency across RMH and LRH. The topics primarily focus on self-harm methods, especially on overdosing with specific medications and dosages, aligning with previous findings that self-poisoning is present in over 50\% of self-harm presentations \cite{witt2023characteristics}. However, the low Jaccard similarity indicates differences in word or phrase choices between hospitals, rather than fundamental differences in patient conditions. Notably, 10 unique topics were identified in RMH, mainly related to psychiatric conditions and overdoses. The absence of these topics in LRH may reflect differences in case mix, transfer pathways, documentation templates, or local triage practices.

\textbf{Future direction \space} Our results suggest several potential directions to enhance model generalisability. First, as self-harm themes remain largely consistent across hospitals, future work could incorporate lexical semantics as predictive features. Second, considering that many unique features related to drugs and their dosages, standardising such alphanumeric combinations into higher-level features might enhance cross-hospital model performance \cite{sikora2024common}. Additionally, normalising triage notes into a standardised format could further help reduce lexical and grammatical variability, ensuring more consistent documentation and better model transferability across hospitals.

\section{Conclusion}
This study compared ED triage notes between RMH and LRH, using selected TF-IDF features and topic modelling. Our analysis shows differences in the importance of selected TF-IDF features, while BERTopic showed high semantic similarity between hospitals but notable lexical differences in topic representation. These differences may contribute to the performance drop in model generalisation, and point to future directions in enhancing cross-hospital model transferability by normalising text to reduce lexical variability.

\section*{Limitations}
% The "Limitations" section (along with, optionally, a section for ethical considerations) may be up to one page and will not count toward the final page limit. 
This study is limited to two Australian hospitals, which constrains the generalisability of our findings. Differences in dataset size and possible discrepancies in annotation practices, even with expert involvement, may have influenced feature representations.
We also note that TF-IDF and BERTopic were selected for comparability and interpretability, but alternative representations may yield different insights. In particular, the implications for contextual embedding models and large language models (LLMs) remain an open question. If the primary source of cross-site variation is lexical rather than thematic, as our BERTopic results suggest, models with greater capacity to generalise across surface-level lexical differences may partially mitigate the portability gap. However, empirical validation on such models is needed before drawing conclusions.

\section*{Ethical Considerations}
Ethical approval was granted by the Melbourne Health Human Research Ethics Committee (HREC; 2017.342). All data were de-identified and are analysed within a secure institutional infrastructure. Due to the sensitivity of the data, our corpus cannot be publicly shared.

Self-harm detection should be approached with particular caution in clinical contexts. In this study, the primary purpose of these models is public health surveillance rather than clinical decision-making. Our models are not intended to replace clinician judgement. We emphasize that predictive models require careful validation prior to deployment to avoid unintended harm, particularly when applied across institutions.

\section*{Acknowledgements}
We thank Jonathan Knott and Owen Connolly for facilitating data acquisition at the Royal Melbourne Hospital and the Latrobe Regional Health. We also thank Hannah Richards and Lu Zhang for their assistance with manual data coding. 

\section*{Funding}
JR is supported by an NHMRC Investigator Grant (2008460) and the University of Melbourne Dame Kate Campbell Fellowship.

% \bibliography{bionlp25_ref}

\appendix

\section{Appendix}
\label{sec:appendix}

\begin{table}[!t]
\small
    \centering
    \begin{tabularx}{\linewidth}{p{1.1cm} X X}
        \hline
        & \textbf{RMH} & \textbf{LRH} \\
        \hline
        \textbf{Unigram} & od, self, resolved, tablets, pain, hanging, sob, depression, superficial, diazepam 
                & pain, overdose, od, suicide, self, tablets, suicidal, police, razor, attempt \\ 
        \textbf{Bigram}  & self inflicted, intentional od, to end, suicidal intent, polypharmacy od, sudden onset, self harm, suicide attempt, od of, hx depression  
                & selling to, od of, self harmed, overdose of, self inflicted, self harm, by gp, to kill, pain on, on palpation \\ 
        \textbf{Trigram} & self inflicted stab, self inflicted lac, self inflicted lacs, with intent to, in attempt to, pmhx depression anxiety, superficial lacs to, superficial cuts to, section 351 self, self harm to  
                & superficial lacs to, superficial cuts to, superficial lacerations to, to left forearm, under section 351, unknown quantity of, self harm lacs, self harm to, pt states took, wants to die \\ 
        \hline
    \end{tabularx}
    \caption{Top 10 selected features in RMH and LRH.}
    \label{tab:fts_overall_comparison}
\end{table}

\begin{figure}[t]
\centering
  \includegraphics[width=\linewidth]{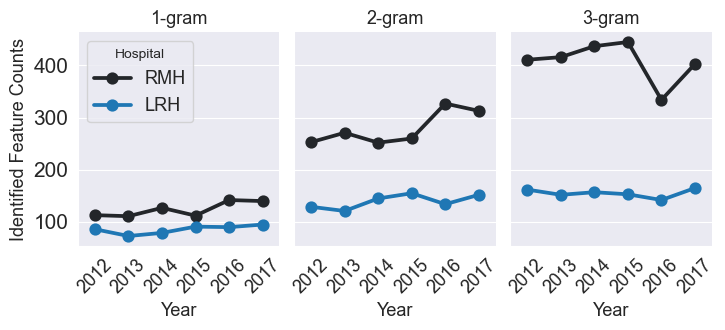}
  \caption{Numbers of selected features each year}
  \label{fig:total_ft_count}
\end{figure}

\end{document}